\documentclass[conference]{IEEEtran}
\IEEEoverridecommandlockouts
\usepackage{cite}
\usepackage{amsmath,amssymb,amsfonts}
\usepackage{algorithmic}
\usepackage{graphicx}
\usepackage{textcomp}
\usepackage{xcolor}
\usepackage{float}
\usepackage{soul} 

\graphicspath{ {./Figures/} }
\def\BibTeX{{\rm B\kern-.05em{\sc i\kern-.025em b}\kern-.08em
    T\kern-.1667em\lower.7ex\hbox{E}\kern-.125emX}}

\begin{document}

\title{A Method for Multi-Robot Asynchronous Trajectory Execution in MoveIt2 (Extended Abstract)}

\author{\IEEEauthorblockN{Pascal Stoop}
\IEEEauthorblockA{\textit{ILT,} \textit{OST}\\
Rapperswil SG, Switzerland \\
pascal.stoop@ost.ch}
\and
\IEEEauthorblockN{Tharaka Ratnayake}
\IEEEauthorblockA{\textit{InIT,} \textit{ZHAW}\\
Winterthur, Switzerland \\
raty@zhaw.ch}
\and
\IEEEauthorblockN{Giovanni Toffetti}
\IEEEauthorblockA{\textit{InIT,} \textit{ZHAW}\\
Winterthur, Switzerland \\
toff@zhaw.ch}
}

\maketitle

\begin{abstract}
This work presents an extension to the MoveIt2 planning library supporting asynchronous 
execution for multi-robot/multi-arm robotic setups. The proposed method introduces a unified way for the execution of both synchronous and asynchronous trajectories by implementing a simple scheduler and guarantees collision-free operation by continuous collision checking while the robots are moving. 

\end{abstract}


\section{Introduction}
When dealing with multi-arm robots or setups with multiple robots where the workspace is shared, there are two main approaches to plan and execute trajectories in a collision free way. Firstly, there is the synchronous approach: All arms (or robots) that are moving at the same time are assumed to be a single moving entity, and traditional planning and execution approaches are being utilized. In the asynchronous approach, trajectories involving the shared workspace are planned independently, so they have to be managed by a central entity to avoid collision. This allows individual arms to operate independently, facilitating the simultaneous management of multiple unrelated tasks by multiple arms \cite{Seyed_2018}. 

By allowing independent motion, asynchronous execution allows for a more efficient utilization of the arms when tasks vary in duration. For example, in a packaging facility, some items need quick packaging, while others require longer, complex packaging. With asynchronous execution, robotic arms can be individually assigned tasks of different duration, avoiding planning and executing combined coordinated motions for different tasks.

While several works address safe operation of multi-arms with synchronous execution
, fewer deal with asynchronous execution, either posing it as a multi-agent problem\cite{Grady_2010},  using trajectory reservation\cite{Planrob}, or using online collision detection as in \cite{GSoC_2022}\footnote{The work presented in this paper was developed before and without knowledge of the cited work. Our implementation is available at: https://github.com/stooppas/moveit2 }. 
%
MoveIt2
\cite{Coleman_moveit} is the most adopted library for robotic arm motion planning and execution, but, at the time of writing, while supporting multi-arm system, it only allows synchronous execution and lacks asynchronous execution. In this work, we report on extending the framework by adding a collision checking approach involving a simple scheduler.


\section{Approach}
MoveIts functionalities are exposed through the \emph{move group} module, which manages and plans motions for specific subsets of joints and links of a robot. The \emph{planning scene} component collects the robot's joint configurations, positions of links, collision objects in the environment and provides collision-checking capabilities.
The \emph{trajectory execution manager} handles the execution of planned trajectories.  Its key responsibilities include trajectory splitting, dispatching these sub-trajectories to their respective controllers and relaying execution information. 
With the current implementation of MoveIt, it is only possible to do synchronized movements, as for safety reason only a single trajectory can be executed at the same time.

In order to support asynchronous execution, we introduced the following changes to MoveIt:
1) Collision detection strategies for multiple trajectories 
2) Central execution scheduler that ensures collision free execution 
3) Online collision detection.


\subsection{Central scheduler}

\begin{figure*}[htbp]
\begin{center}
\includegraphics[width=0.8\textwidth]{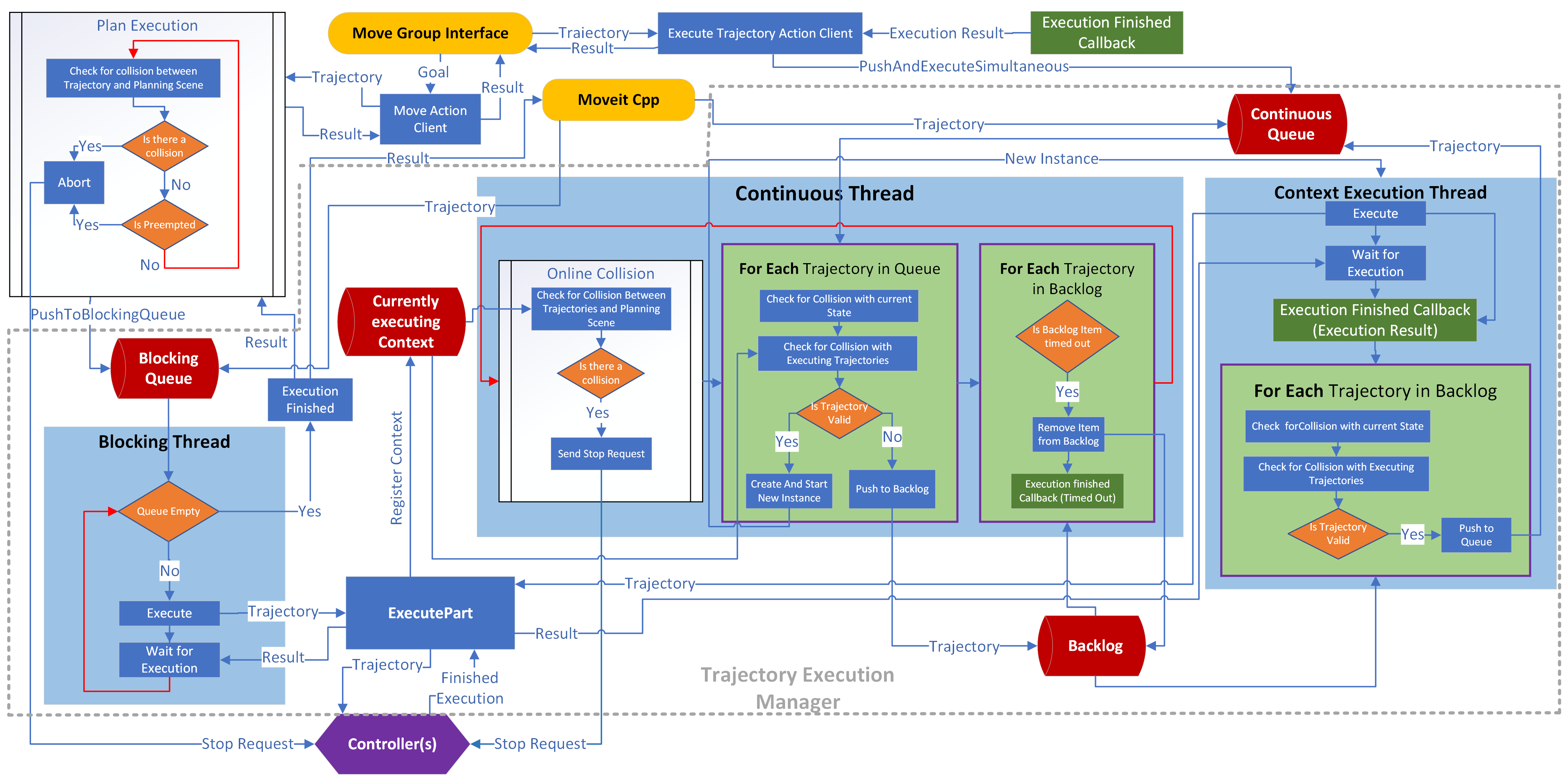}
\caption{Proposed Trajectory Execution Manager.}
\label{fig:exec_mgr}
\end{center}
\end{figure*}

To allow the trajectory execution manager to handle multiple trajectories, the blocking execution that was previously present has been replaced with a continuous execution queue (top right in Figure \ref{fig:exec_mgr}). Any new trajectory $t_n$ added to the queue will be collision checked against all currently running trajectories $t_r \in T$. 
If no collision is expected, $t_n$ will be scheduled for immediate execution and the execution manager will store its information for future collision avoidance. 
Instead, in case of expected collision, $t_n$ is added to a backlog queue (bottom right in figure), similarly to the implementation suggested by Felix von Drigalski in \cite{Felix_2020}. 
As soon as the running trajectory $t_r$ that was causing a collision with $t_n$ is finished, $t_n$ will be moved from the backlog into the continuous queue and checked against collisions to be executed again.
To avoid deadlocks and minimize execution duration, all trajectories are assigned a backlog timeout. If the execution manager is not able to schedule a trajectory within said timeout, the trajectory execution aborts, allowing the user to re-plan. 

\subsection{Collision Detection}

Collision avoidance between a new and a running trajectory can be addressed by ``sweeping'' methods that reserve all the space occupied by an entire trajectory in the workspace (as done in e.g., \cite{Planrob}). However, this reduces the motion space available for planning. For this reason, we chose to check for collision in a time dependent manner. 

The default collision checking library in MoveIt is FCL (Flexible Collision Library), which allows for discrete collision detection for a single robot state. To be able to check for collision in a time-variant manner, we will have to discretize the trajectory and check for every discrete time step. At first we tried evaluating collisions between a trajectory's discrete robot state at each timestamp and the interpolated state of another trajectory at the corresponding timestamp, but this led to undetected collisions, so we resorted to use a configurable parameter for the time step used to break trajectories into discrete steps. This allows to tune the trade-off between performance and security.


This approach is still of course sub-optimal,  as it won't detect collisions between discrete time-steps. To solve this issue, Continuous Collision Detection (CCD) could be used in a future implementation. 

\subsection{Online collision detection}

MoveIt incorporates real-time trajectory validity assessment using collision checking for single trajectories, yet collision checks only consider a single trajectory as well as the continuously updated planning scene. This design, meant to account for a single executing trajectory, may incur in a high executon overhead due to extensive collision evaluations. A more efficient approach is sought to accommodate multiple trajectories while expanding collision checks between executing trajectories and the planning scene. Our proposed solution consolidates all robot states from concurrent trajectories at discrete intervals, subjecting this combined state to self-collision checks and evaluations against the planning scene. Discretization strategies to balance efficiency and safety vary, with the preferred approach involving periodic collision checks with planning scene updates.


\section{Conclusions and Future Work}
\begin{figure}[!htb]\vspace*{-0.3cm}
\centerline{
\includegraphics[scale=0.063]{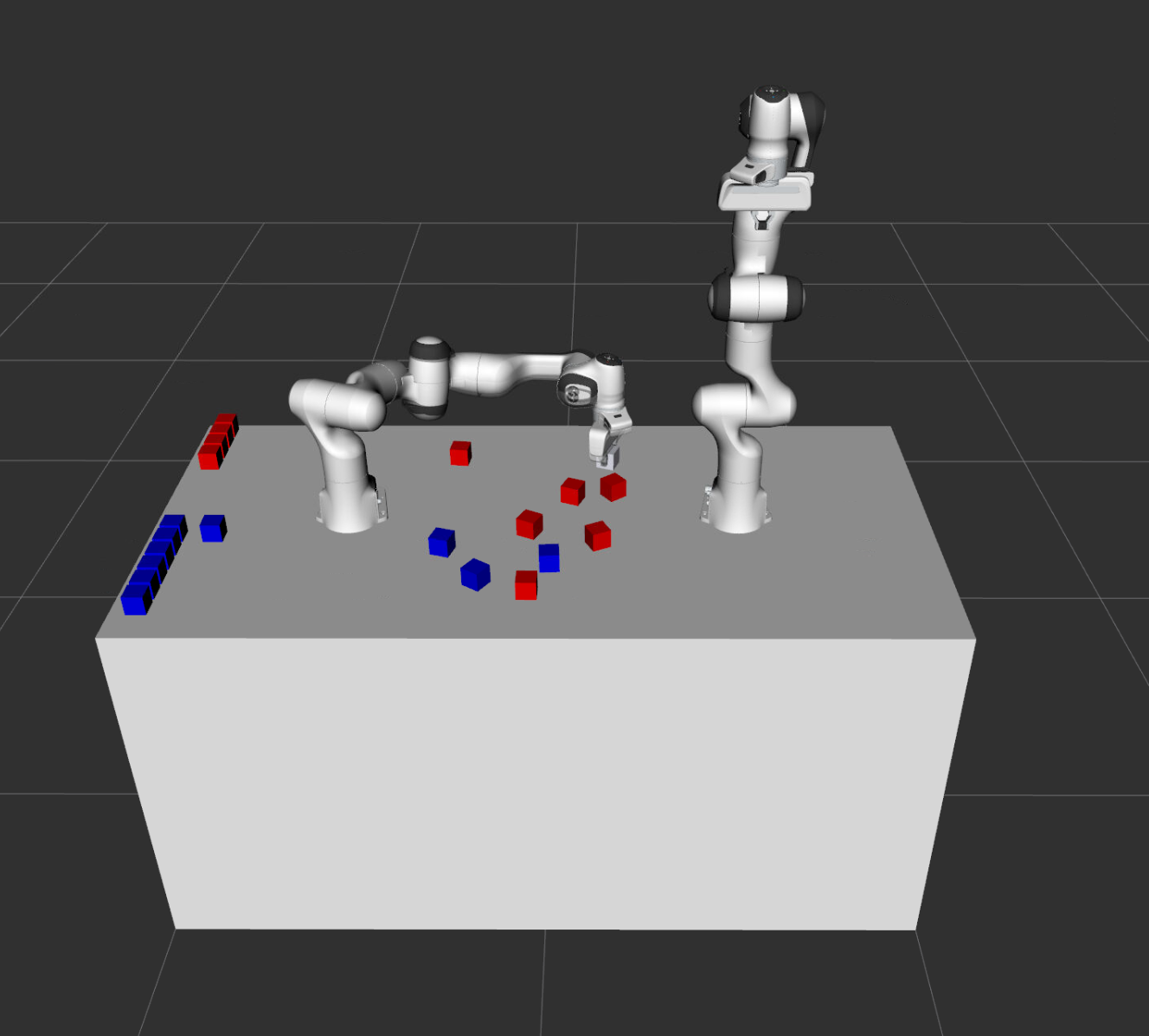}
\includegraphics[scale=0.063]{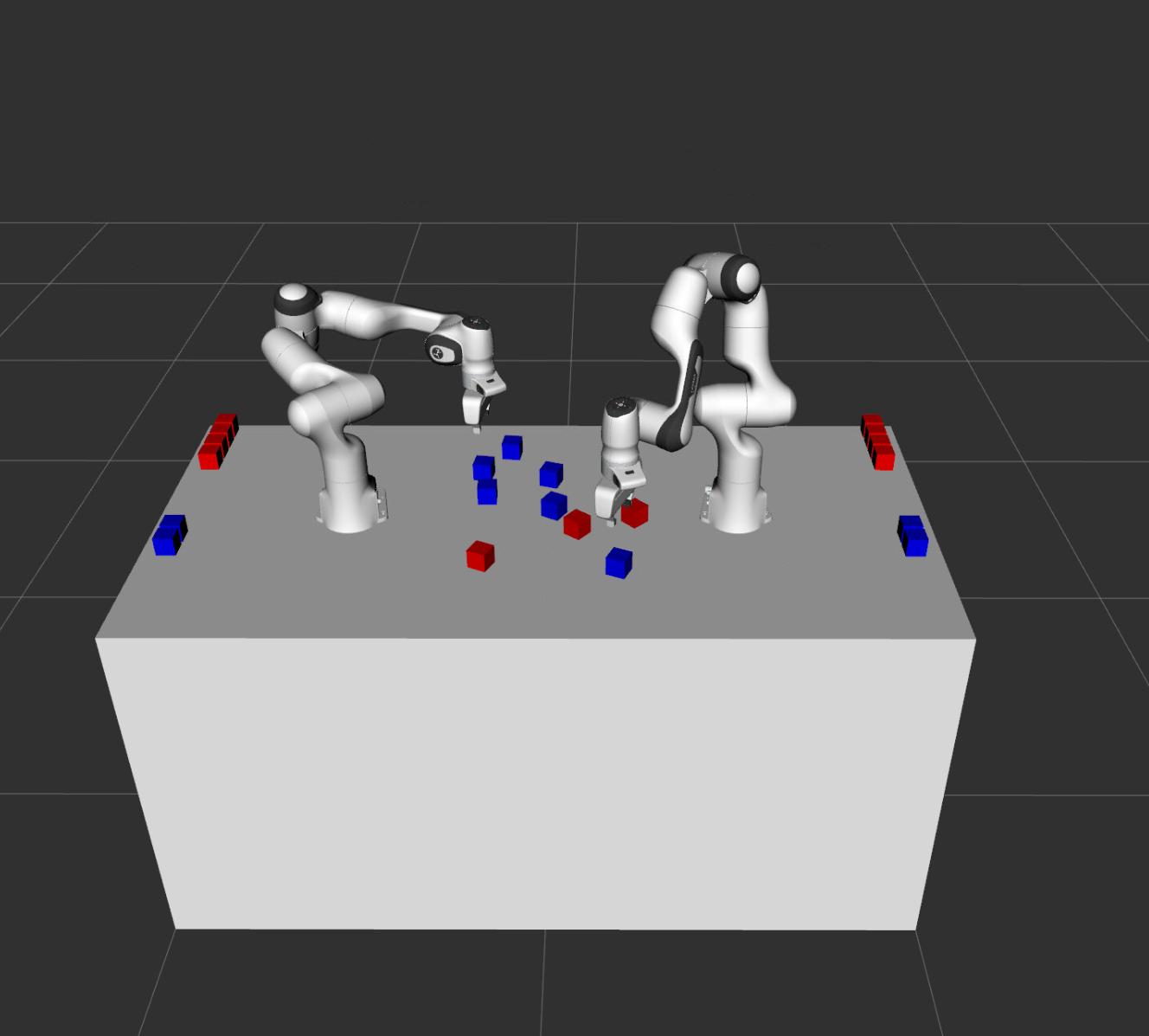}
\includegraphics[scale=0.063]{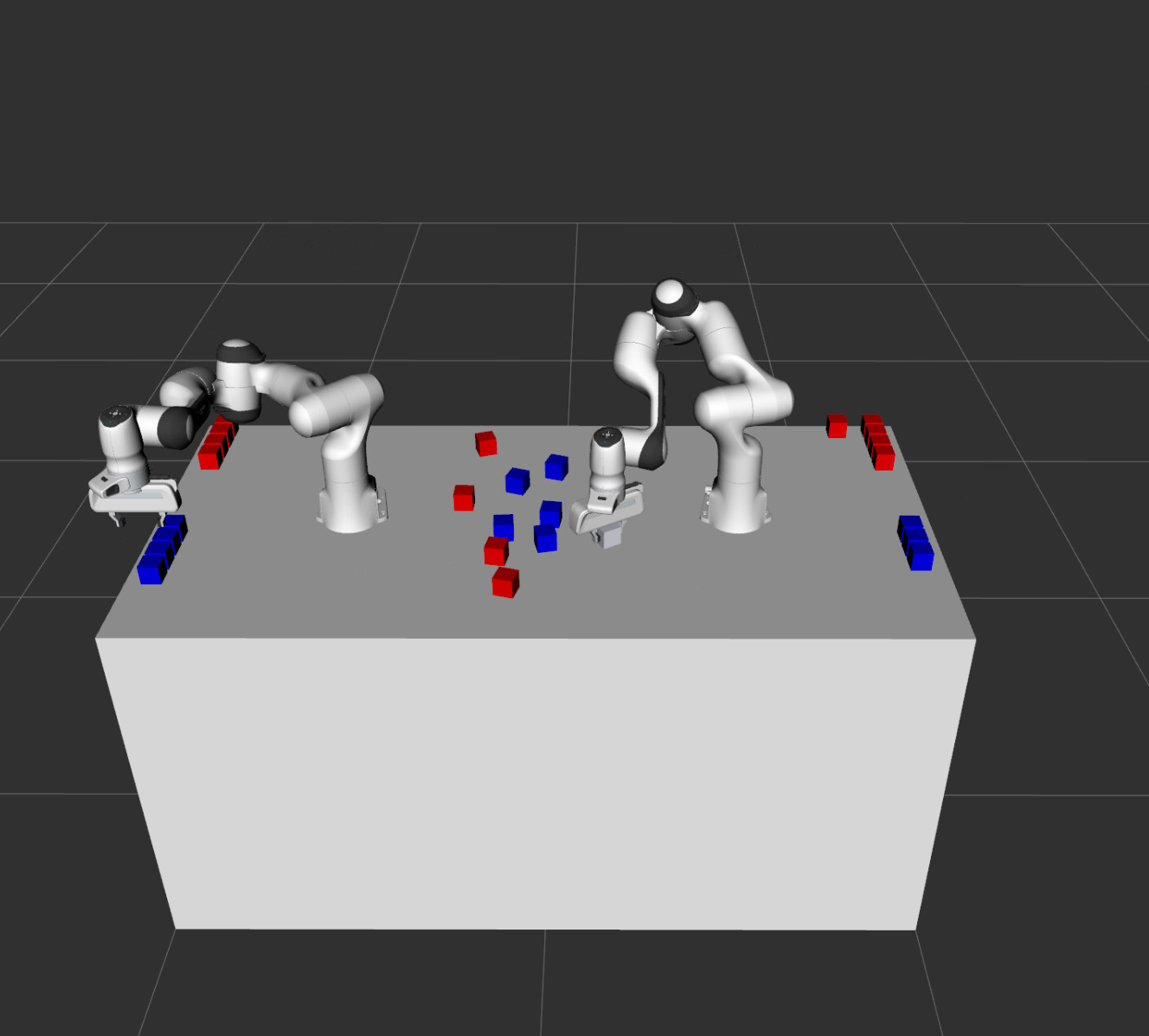}
}
\caption{baseline with only one arm \textit{[left]}, two arms synchronous movements \textit{[middle]} and two arms asynchronous movements \textit{[right]} }
\label{fig:platforms}
\end{figure}

In this paper we introduced an extension to allow MoveIt2 to support multi-arm / multi-robot asynchronous trajectory execution.
Through a simulated pick and place scenario of two 7 DoF Panda arms in a shared workspace (see Figure \ref{fig:platforms}), we are currently running experiments that are 1) demonstrating the correct functioning of the implementation and collision avoidance logic; 2) proving the advantage of this approach with respect to synchronous execution for specific workloads; and 3) measuring the overhead induced by the backlog and re-planning logic.


\vspace{-1em}

\bibliographystyle{IEEEtran}
\bibliography{reference.bib}

\begin{thebibliography}{1}
\providecommand{\url}[1]{#1}
\csname url@rmstyle\endcsname
\providecommand{\newblock}{\relax}
\providecommand{\bibinfo}[2]{#2}
\providecommand\BIBentrySTDinterwordspacing{\spaceskip=0pt\relax}
\providecommand\BIBentryALTinterwordstretchfactor{4}
\providecommand\BIBentryALTinterwordspacing{\spaceskip=\fontdimen2\font plus
\BIBentryALTinterwordstretchfactor\fontdimen3\font minus
  \fontdimen4\font\relax}
\providecommand\BIBforeignlanguage[2]{{%
\expandafter\ifx\csname l@#1\endcsname\relax
\typeout{** WARNING: IEEEtran.bst: No hyphenation pattern has been}%
\typeout{** loaded for the language `#1'. Using the pattern for}%
\typeout{** the default language instead.}%
\else
\language=\csname l@#1\endcsname
\fi
#2}}

\bibitem{Seyed_2018}
{Mirrazavi Salehian et al.}, ``A unified framework for coordinated multi-arm
  motion planning,'' \emph{The International Journal of Robotics Research},
  vol.~37, no.~10, pp. 1205--1232, 2018.

\bibitem{Grady_2010}
{D. Grady et al.}, ``Asynchronous distributed motion planning with safety
  guarantees under second-order dynamics,'' vol.~68, 01 2010, pp. 53--70.

\bibitem{Planrob}
C.~A. Meehan, M.~Roberts, and L.~M. Hiatt, ``{Asynchronous Motion Planning and
  Execution for a Dual-Arm Robot},'' \emph{ICAPS}, 2022.

\bibitem{GSoC_2022}
\BIBentryALTinterwordspacing
(2022) Gsoc 2022: Simultaneous trajectory execution. [Online]. Available:
  \url{https://github.com/ros-planning/moveit/issues/3156}
\BIBentrySTDinterwordspacing

\bibitem{Coleman_moveit}
{D. Coleman et al.}, ``Reducing the barrier to entry of complex robotic
  software: a moveit! case study,'' \emph{CoRR}, vol. abs/1404.3785, 2014.

\bibitem{Felix_2020}
F.~von Drigalski~et al., ``Robots assembling machines: learning from the world
  robot summit 2018 assembly challenge,'' \emph{Advanced Robotics}, vol.~34,
  2020.

\end{thebibliography}

\end{document}